\documentclass[letterpaper]{article}
\usepackage{iccc}

\usepackage{times}
\usepackage{helvet}
\usepackage{courier}
\usepackage[utf8]{inputenc}
\usepackage{microtype}
\usepackage{graphicx}
\usepackage{subcaption}     
\usepackage{booktabs}       
\usepackage{hyperref}
\usepackage{adjustbox}      
\usepackage{todonotes}
\usepackage{xcolor}
\usepackage{stfloats}       

\usepackage{amsfonts}
\usepackage{mathtools}
\usepackage{makecell}
\usepackage{placeins}       
\usepackage{enumitem}       

\title{Simultaneous Multiple-Prompt Guided Generation \\
Using Differentiable Optimal Transport}

\author{Yingtao Tian\\
Google Brain\\
Tokyo, Japan
\And
Marco Cuturi \\
Google Brain (currently at Apple)\\
Paris, France\\
\And
David Ha \\
Google Brain\\
Tokyo, Japan
}

\setlist{leftmargin=0.0mm,label={}}

\newcommand{\removespacetop}{\vspace*{-0.0cm}}
\newcommand{\removespaceovercaption}{\vspace*{-0.0cm}}
\newcommand{\removespacebottom}{\vspace*{-0.0cm}}

\newcommand\defeq{\mathrel{\overset{\makebox[0pt]{\mbox{\normalfont\tiny\sffamily def}}}{=}}}
\DeclareMathOperator*{\argmin}{arg~min} 

\begin{document}

\maketitle









\begin{figure*}[!hb]
    \begin{center}
    \removespacetop
    \begin{subfigure}[t]{0.52\textwidth}
        \includegraphics[width=0.49\textwidth]{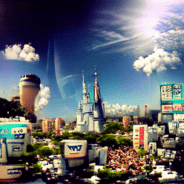}
        \hspace*{\fill}
        \includegraphics[width=0.49\textwidth]{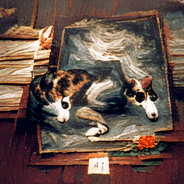}
        \caption{Generated images from \emph{two-prompts} using our method. (Left) ``\emph{Walt Disney World.}'' and ``\emph{a daytime picture of Tokyo.}'' (Right) ``\emph{A painting of cat.}'' and ``\emph{A painting of dog.}''.}
    \end{subfigure}
    \hspace*{\fill}
    \begin{subfigure}[t]{0.42\textwidth}
        \includegraphics[width=\textwidth]{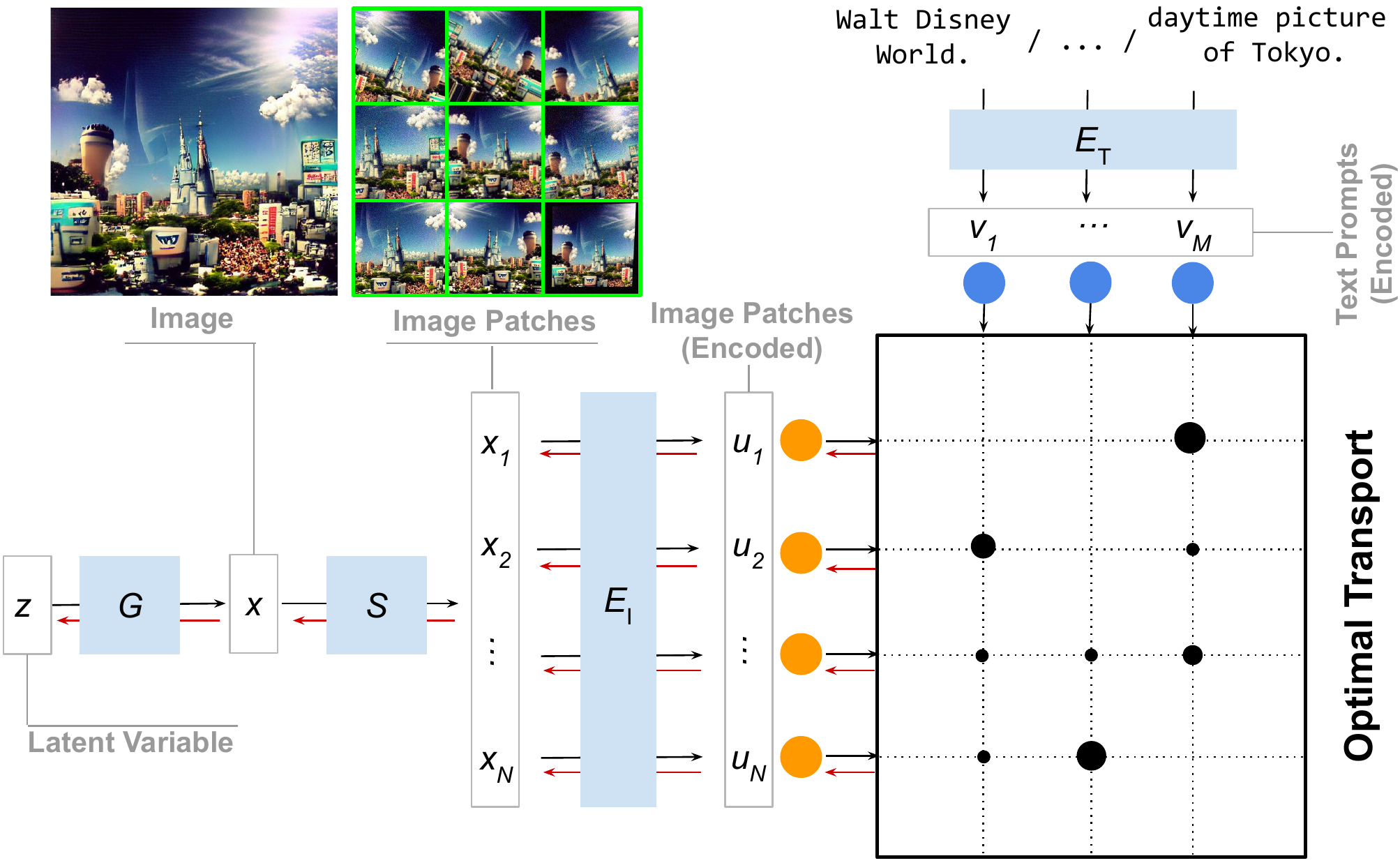}
        \caption{The architecture of our work. Iteratively, the loss is computed forward (marked by $\rightarrow$) and the gradient is calculated backward (marked by $\color{red} \leftarrow$) to update the latent variable $z$.}
    \end{subfigure}
    \removespaceovercaption
    \caption{Our method illustrated with generated images and the architecture.
    In contrast, the existing method would fail with these two-prompts, producing images with less diverse features (left) or a painting with much different art style than single prompt (right). 
    This is because the existing method of taking the mean cannot treat different parts of the image separately, and vector arithmetic in the latent space introduces uncontrollable changes in the semantics. 
    Detailed analysis can be found in text.
    All figures in this paper are \emph{generated} using pre-trained CLIP and VQGAN models, both publicly released under MIT license.
    }
    \end{center}
\end{figure*}

\begin{abstract}
    Recent advances in deep learning, such as powerful generative models and joint text-image embeddings, have provided the computational creativity community with new tools, opening new perspectives for artistic pursuits. \textit{Text-to-image synthesis} approaches that operate by generating images from text cues provide a case in point. These images are generated with a latent vector that is progressively refined to agree with text cues. To do so, patches are sampled within the generated image, and compared with the text prompts in the common text-image embedding space; The latent vector is then updated, using gradient descent, to reduce the mean (average) distance between these patches and text cues.
    While this approach provides artists with ample freedom to customize the overall appearance of images, through their choice in generative models, the reliance on a simple criterion (mean of distances) often causes mode collapse: The entire image is drawn to the \textit{average} of all text cues, thereby losing their diversity.
    To address this issue, we propose using matching techniques found in the optimal transport (OT) literature, resulting in images that are able to reflect faithfully a wide diversity of prompts. We provide numerous illustrations showing that OT avoids some of the pitfalls arising from estimating vectors with mean distances, and demonstrate the capacity of our proposed method to perform better in experiments, qualitatively and quantitatively.
\end{abstract}
\section{Introduction}

The computational creativity community has been at the forefront of engaging with recent advances in deep learning, adopting early on generative models that are able to produce high-quality text and images. Such models offer varying degrees of realism and control to the artist, enabling the generation of results with artistic value.
Recent advances have brought forward models that can produce images from natural language prompts, using pre-trained image generative models guided by text descriptions~\cite{radford2021learning}.
The computational creativity community has seized this opportunity, has shared large bodies of code~\cite{samburtonking2021introduction,wang2021bigsleep} and generated a large body of artwork, some of which has been curated online~\cite{snell2021,murdock}.

These tools are favoured by artists because they can shape generation in various ways: For instance, a relevant generative model can be used  in that the style of pieces of art that can be produced can be efficiently guided by selecting a relevant generative model $G$.  While this degree of freedom is useful, little has changed on how text prompts are handled in that pipeline:

Images can be generated using the following pipeline: The user supplies a generative model $G$ and a text prompt $t$. An initial latent vector $z$ is sampled randomly; the fit between its corresponding image $x=G(z)$ and the desired prompt $t$ is quantified using their distance in a common CLIP embedding space; In order to minimize that distance, $z$ is updated iteratively using gradient steps. Because both the CLIP embedding and $G$ are differentiable, gradients for these distances are obtained using automatic differentiation.
 
In practice, a few more tricks are needed to produce convincing images. To accommodate the important artistic requirement that multiple concepts appear in images, several text prompts are allowed, but are pre-aggregated in embedding space to result in a composite prompt vector. Next, rather than consider the entire image against that composite prompt vector, several patches with random size, orientation and placement are sampled within the image $x$, and are then compared with the composite prompt, before these distances are averaged to form the overall loss.

These tricks rely therefore on aggregations: the mean of various prompt embeddings is used to define a single target prompt, and the various distances of all patches to that target are also reduced to their average.
We argue, and we show later in the paper, that this reliance on averages can cause several issues, causing notably generated images to have parts that are uniformly closer to all prompts, thus defeating the original motivation of using multiple prompts to obtain artistic images with diverse objects. Another important drawback of averaging prompt embeddings is that it can potentially introduce uncontrollable changes in semantics, with a mean prompt embedding falling in a region of the embedding space with no corresponding meaning.

We propose to address this issue by treating the embedded patches of generated image and texts as vectors sampled from two probability distributions, and to use computational optimal transport (OT)~\cite{peyre2019computational} to find the best matching between them.
As its name suggests, OT tries to find the minimal total effort required to ``move'' all patches towards texts, using the pairwise distance as the cost for measuring said effort. OT brings two advantages over simply taking the mean:
(1) Since patches are randomly sampled, it encourages the intrinsic diversity \emph{inside} a single generated image.
(2) OT does not involve vector arithmetic in the latent space, sidestepping issues that may arise from the non-existing semantic of a mean prompt vector.
Concretely, we use Sinkhorn's Algorithm~\cite{cuturi2013sinkhorn,sejourne2019sinkhorn} for the matching,
in a way that is efficient and, most importantly, differentiable using OTT-JAX~\cite{cuturi2021ott}.
Such differentiability is crucial to allow the computation of gradient all the way back to $z$.
 
Bringing all pieces together, our proposed use of OT enables the generation of images that are diverse and without the issue of unwanted extra semantics, as demonstrated empirically in the paper.
Furthermore, since our proposed method only changes how pairs of (patch, prompt) distances are recombined, it is orthogonal to other existing parts of the pipeline, and consists, implementation-wise, in a simple drop-in replacement of mean operations by optimised matchings (incidentally, taking means can be interpreted as the most naive approach conceivable to match pairs). We start this paper with a background section, needed to detail next our methodology, which is illustrated and validated in various experiments that showcase its performance, and explain why it is able to solve several issues arising from an over-reliance on mean distances and mean prompt embeddings.

\section{Background}

In this section, we review two pillars of our work, \emph{prompt-guided image generation} and \emph{differentiable optimal transport}.
We argue in this paper that combining both is crucial to address issues we observe in existing generation methods.

\subsection{Prompt Guided Image Generation}
A notable trend in the field of computational creativity is to guide image generation using natural language as prompts.
These \textit{text-to-painting} synthesis tools allow artists to specify the content of a painting using prompts from natural languages.
This text-driven generation has revolutionized the computational generation of artworks, as evidenced in online curated collections~\cite{snell2021,murdock}.
These advances are made possible by combining two innovations from deep learning:

\begin{itemize}
    \item \emph{Powerful image generative models}.
    Such models include recent generative adversarial networks (GANs)~\cite{karras2019style,karras2020analyzing,karras2021alias}, variational autoencoders~\cite{van2017neural} and diffusion models~\cite{ho2020denoising,song2020denoising,nichol2021improved,dhariwal2021diffusion}, that can produce images with high fidelity and diversity.
    Formally, this process can be denoted as $x=G(z)$ where the generative model $G: \mathbb{R}^{d_z} \to \mathbb{R}^{h \times w \times 3}$ converts a latent space variable $z \in \mathbb{R}^{d_z}$ to an RGB image of height $h$, weight $w$ and $3$ color channels. $x \in \mathbb{R}^{h \times w \times 3}$.
    $z$ could be further manipulated to allow generating more suitable $x$~\cite{li2021surrogate}, allowing artist to control the generation of artworks that fall in desired genres~\cite{jin2017towards}. 
    
    \item \emph{Joint modeling of images and natural language}.
    This idea has been long in the making~\cite{thomee2016yfcc100m,li2017learning}, but only recently given a convincing  implementation thanks to progress in natural languages modeling~\cite{raffel2019exploring,brown2020language}, and notably the ability to embed jointly images and text so well that the need for task-specific fine-tuning is eliminated, as shown in CLIP~\cite{radford2021learning}.
    CLIP provides two jointly-trained, differentiable encoders, $E_{\mathrm{I}}: \mathbb{R}^{h \times w \times 3} \to \mathbb{R}^d$ and $E_{\mathrm{T}}: \mathcal{T} \to \mathbb{R}^d$, for image and text respectively.
    We do not further elaborate the domain of text $\mathcal{T}$ as it is not the focus of this work.
    Formally, given an image $x$ and a text $t$, and a distance function
    $D: \mathbb{R}^{d} \times \mathbb{R}^d \to \mathbb{R}_{+}$ the encoded image $u=E_{\mathrm{I}}(x)$ and the encoded text $v=E_{\mathrm{T}}(t)$ are in a common comparable space $\mathcal{U} = \mathbb{R}^d$, and $D(u, v)$ measures the similarity between $x$ and $t$.
    In the case of CLIP that is trained with cosine distance,
    practically $D$ could be chosen as cosine distance or geodesic distance, both effectively measuring the angle between the two vectors and being trivially differentiable.
    Ideally, text-driven image generation is now feasible by iteratively adjusting the latent space vector $z$,
    to minimize $D(u, v)$, the distance between the encoded image $x=G(z)$ and encoded user-specific prompt $t$.
    As $G$, $E_{\mathrm{I}}$ and $D$ are differentiable, $z$ could be updated using gradient Descent:
    $z \gets z - \gamma \nabla_{z} F(z)$
    where $\nabla_{z} F$ is the gradient of $F$ defined as $F(z) = D(E_{\mathrm{I}}(G(z)), E_{\mathrm{T}}(t))$ and $\gamma$ is a learning rate.
\end{itemize}
     
Using a distance from a single image to a single prompt is usually too restrictive. Therefore, and in practice, the distance is computed over pairs of multiple images and texts as follows:
On the image side, $n$ patches (a.k.a.~cutouts. We use these two terms interchangeably), 
which we denote as $x_1, \cdots, x_n = S(x)$ are randomly sampled from image $x$ in the fashion of image data augmentation~\cite{shorten2019survey}.
We assume $x_i \in \mathbb{R}^{h \times w \times 3}$ still holds since we can trivially add a resizing step at the end of augmentation.
This practice serves as a regularizer to ensure numerical stability and avoid fitting into regions of $z$ where $G$ has bad support.
On the text side, $m$ text prompts, denoted as $t_{1}, \cdots, t_{m}$, are often considered,
which allows artists to explore the possibilities of art by combining multiple texts as directions.
Again, they are encoded accordingly, giving $u_1, \cdots, u_n: u_i = E_{\mathrm{I}}(x_i) \in \mathbb{R}^d$ and $v_1, \cdots, v_m: v_j = E_{\mathrm{T}}(t_j) \in \mathbb{R}^d$.
These pairwise distances are then combined to form a loss, which is
\begin{equation} \label{eq:f_mean}
F(z) = \mathrm{Mean}_{D}(z) \defeq \frac{1}{mn}  \sum_{1\leq i\leq n, 1\leq j \leq m} D(u_i, v_j),
\end{equation}
and thus the gradient $\nabla_z F$ reads
\begin{equation} \label{eq:nabla_f_mean}
\nabla_z F = \left(
  \sum_{1\leq i \leq n}
  \frac{\partial \mathrm{Mean}_{D}}{\partial u_i}
  \frac{\partial u_i}{\partial x_i}
  \frac{\partial x_i}{\partial x}
  \right)
  \frac{\partial x}{\partial z}
\end{equation}
where
\begin{align}
\begin{split} \label{eq:partial_terms}
\frac{\partial \mathrm{Mean}_{D}}{\partial u_i} = \frac{1}{nm} \sum_{1\leq j\leq m} \frac{\partial D(u_i, v_j)}{\partial u_i} \\
\frac{\partial u_i}{\partial x_i} = \nabla_{x} E_{\mathrm{I}}(x_i), \quad
\frac{\partial x}{\partial z} = \nabla_{z} G(z)
\end{split}
\end{align}
and $\partial x_i / \partial x$ is defined as long as the random sampling is differentiable w.r.t.\ the input image $x$ which is often the case of data augmentations.
This framing of text-driven generation has been applied to different generators $G$, yielding a variety of artistic results: using unconditional GAN generation, like BigGAN~\cite{wang2021bigsleep}, VQGAN~\cite{samburtonking2021introduction} and SIREN~\cite{wang2021deepdaze};
conditional generation using GAN, such as StyleCLIP~\cite{patashnik2021styleclip}, that enables editing existing images.
In addition to GANs, it can also be applied to Diffusion models~\cite{crowson2021clip,kim2021diffusionclip,nichol2021glide}.

\subsection{Differentiable Optimal Transport}

Optimal transport (OT), as its name suggests, can be understood as finding an efficient way to `move’ or `transport', the mass from a probability distribution to another distribution.
We borrow notations from the survey book~\cite{peyre2019computational} and focus on one of the canonical OT formulations, one that was proposed in~\cite{kantorovich1942transfer}.
A discrete measure with weights $a$ on locations $u_1, \cdots, u_n$ would be denoted as $\alpha = \sum_{1\leq i \leq n} a_i \delta_{u_i}$, where notation $\delta_u$ stands for a Dirac mass at location $u$.
Similarly, for weights $b$ on locations $v_1, \cdots, v_m$ we have  $\beta = \sum_{1\leq j \leq m} b_j \delta_{v_j}$.
A possible way to map a discrete measure $\alpha$ onto $\beta$, given a cost matrix $\mathbf{C} \in \mathbb{R}_{+}^{n\times m}$,
can be represented with a coupling matrix $\mathbf{P} \in \mathbb{R}_{+}^{n\times m}$, where the amount of mass transported from the $i$-th location in $\alpha$ to $j$-th location in $\beta$ is stored as $\mathbf{P}_{i,j}$.
The set of admissible couplings, $\mathbf{U}$, is defined through $a$ and $b$ as
\[
\mathbf{U}(a, b) \defeq \left\{ \mathbf{P} \in \mathbf{R}_{+}^{n\times m}: \sum_j \mathbf{P}_{i,j} = a, \quad \sum_i \mathbf{P}_{i,j} = b \right\},
\]
These row- and and column-sum constraints for $P$ indicate that the entire mass from $\alpha$ is indeed transported to $\beta$.
Kantorovich's problem of interest is
\[
\mathrm{L}(a, b, \mathbf{C}) \defeq \min_{\mathbf{P} \in \mathbf{U}(a, b)} \langle \mathbf{C}, \mathbf{P} \rangle \defeq \sum_{i,j} \mathbf{C}_{i,j} \mathbf{P}_{i,j},
\]
which can be solved using linear programming, notably network flow solvers.
The linear programming route, while well established, has a few drawbacks: it is slow, with an unstable solution. A possible workaround is to add an entropic regularization term,
where the entropy of $P$ reads $\mathbf{H}(\mathbf{P}) \defeq -\sum_{i,j} \mathbf(P)_{i,j} ( \log (\mathbf{P}_{i,j})-1).$
The regularized problem reads:
\[
\mathrm{L}^{\epsilon} (a, b, \mathbf{C}) \defeq \min_{\mathbf{P} \in \mathbf{U}(a, b)} \langle \mathbf{C}, \mathbf{P} \rangle - \epsilon  \mathbf{H}(\mathbf{P}).
\]
This regularization has several practical virtues: the regularized problem can be solved efficiently with Sinkhorn's Algorithm~\cite{cuturi2013sinkhorn,sejourne2019sinkhorn}, a fast iterative algorithm that only uses matrix-vector arithmetic.
Another advantage, equally important in our setting, is that this approach, as implemented in OTT-JAX~\cite{cuturi2021ott} results in fully differentiable quantities.
Namely, assume that the cost matrix $\mathbf{C}$ is provided in the form of a differentiable function resulting in entries $\mathbf{C}_{i,j} \defeq \mathbf{C}(u_i, v_j)$. Then the gradient of $\mathrm{L}^{\epsilon}$ w.r.t. $u_i$ exists and is defined everywhere:
\begin{equation} \label{eq:partial_lc}
\forall i, 1 \leq i \leq n, \left\|\frac{\partial \mathrm{L}^{\epsilon}}{\partial u_i}
\right\|<\infty.
\end{equation}
Not that the optimal solution $P^\epsilon$ corresponding to $\mathrm{L}^{\epsilon}$ can also be differentiated w.r.t. any of the relevant inputs, using the implicit function Theorem~\cite{krantz2002implicit}, as proposed in OTT-JAX~\cite{cuturi2021ott}, but this is not used in this paper because we rely on Danskin's Theorem~\cite{danskin1966theory} (a.k.a Envelope Theorem) to differentiate $\mathrm{L}^{\epsilon}$ w.r.t. $\mathbf{C}$.

\begin{figure*}[!hbt]
    \centering
    \newcommand\subwidth{0.35\textwidth}
    \hspace*{\fill}
    \begin{subfigure}[t]{\subwidth}
        \includegraphics[width=\textwidth]{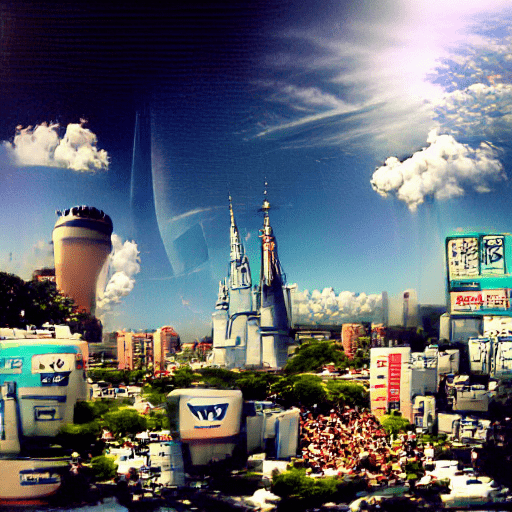}
        \caption{Our Proposed Method (OT)}
    \end{subfigure}\hfill
    \begin{subfigure}[t]{\subwidth}
        \includegraphics[width=\textwidth]{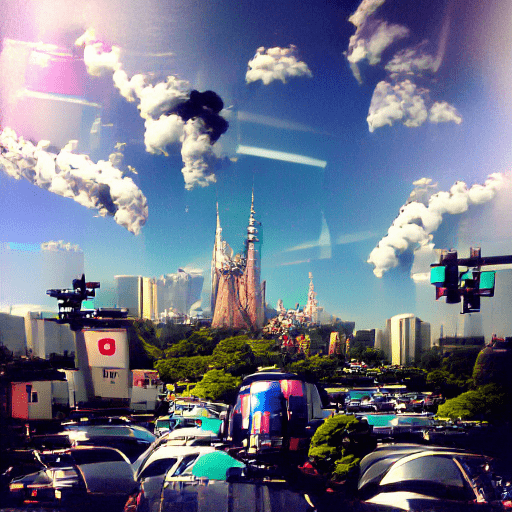}
        \caption{Baseline (Mean)}
    \end{subfigure}
    \hspace*{\fill} 
    \removespaceovercaption
    \caption{The generated image from two prompts: ``Walt Disney World.'' and ``daytime picture of Tokyo.'' Compare with the baseline, our methods generates images with better diversity (Disney-like architecures vs. city scense) while blending them well.}
    \label{fig:two-prompts-generated-images}
    \removespacebottom
\end{figure*}
\begin{figure*}[!thb]
    \centering
    
    \newcommand\subwidth{0.35\textwidth}
    
    \hspace*{\fill}
    \begin{subfigure}[t]{\subwidth}
        \includegraphics[width=\textwidth]{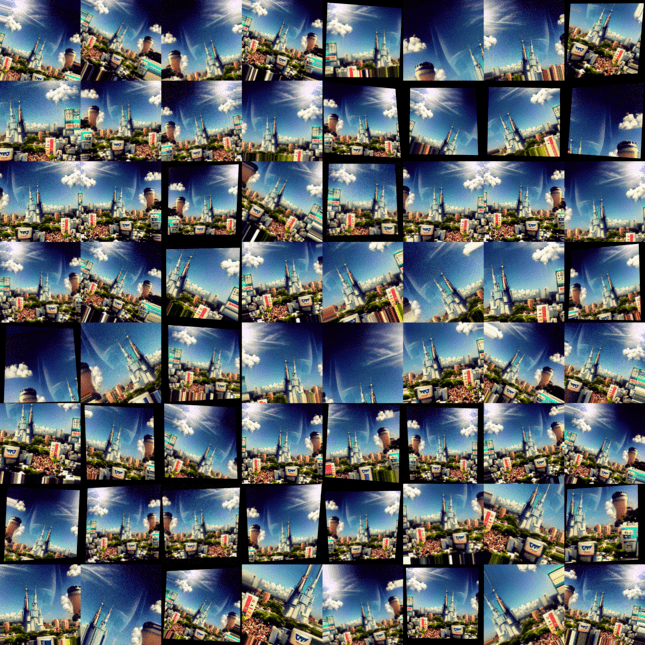}
        \caption{Patches (cutouts) from our method (OT)}
    \end{subfigure}\hfill
    \begin{subfigure}[t]{\subwidth}
        \includegraphics[width=\textwidth]{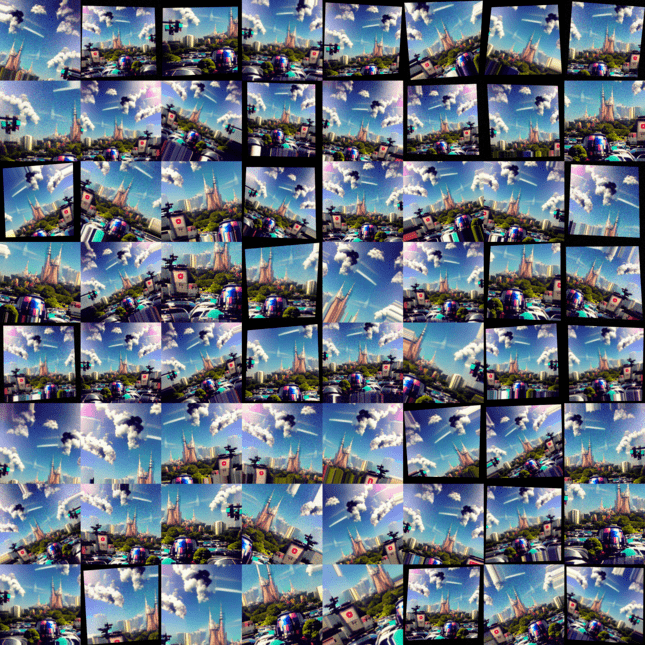}
        \caption{Patches (cutouts) from  baseline (Mean)}
    \end{subfigure}
    \hspace*{\fill}

    \vspace{0.2cm}

    \newcommand\imgwidth{8cm}

    \begin{subfigure}[b]{\textwidth}
        \centering
        \begin{adjustbox}{width=0.95\textwidth}
            \begin{tabular}{lp{\imgwidth}}
                \toprule
                    \makecell[l]{
                    Prompt 0 : Walt Disney World. \\
                    \textbf{36} out of 64 cutouts are closer to Prompt 0.
                    } &
                    \parbox[c]{2em}{\includegraphics[width=\imgwidth]{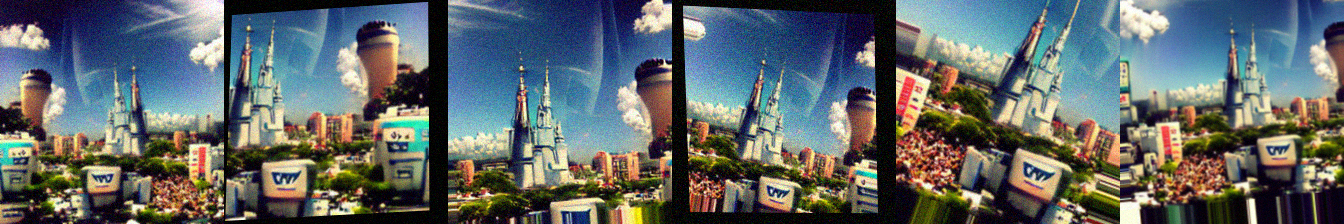}} \\
                \midrule
                    \makecell[l]{
                        Prompt 1 : A daytime picture of Tokyo. \\
                        \textbf{28} out of 64 cutouts are closer to Prompt 1.
                    } &
                    \parbox[c]{2em}{\includegraphics[width=\imgwidth]{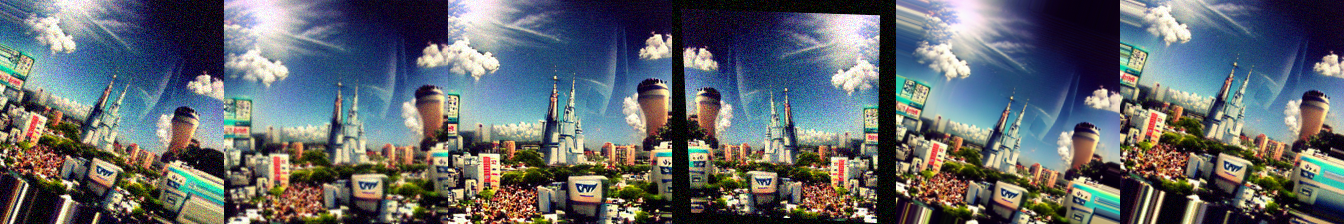}} \\
                \bottomrule
            \end{tabular}
        \end{adjustbox}
        \caption{Prompts closer to each prompt, in our proposed method (OT)}
    \end{subfigure}\hfill

    \begin{subfigure}[b]{\textwidth}
        \centering
        \begin{adjustbox}{width=0.95\textwidth}
            \begin{tabular}{lp{\imgwidth}}
                \toprule
                    \makecell[l]{
                    Prompt 0 : Walt Disney World. \\
                    \textbf{49} out of 64 cutouts are closer to Prompt 0.
                    } 
                    & \parbox[c]{2em}{\includegraphics[width=\imgwidth]{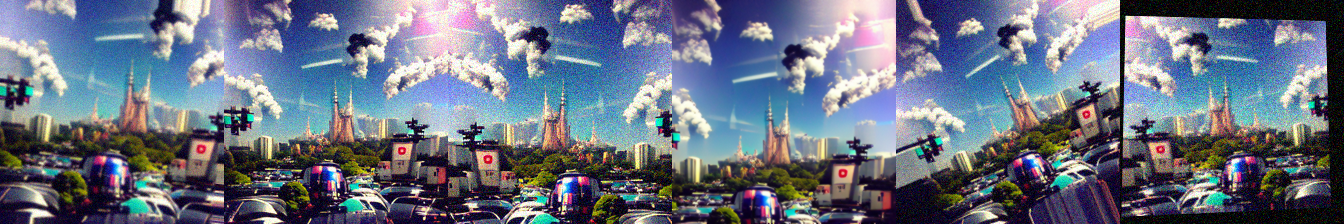}} \\
                \midrule
                \makecell[l]{
                    Prompt 1 : A daytime picture of Tokyo. \\
                    \textbf{15} out of 64 cutouts are closer to Prompt 1.
                } &
                \parbox[c]{2em}{\includegraphics[width=\imgwidth]{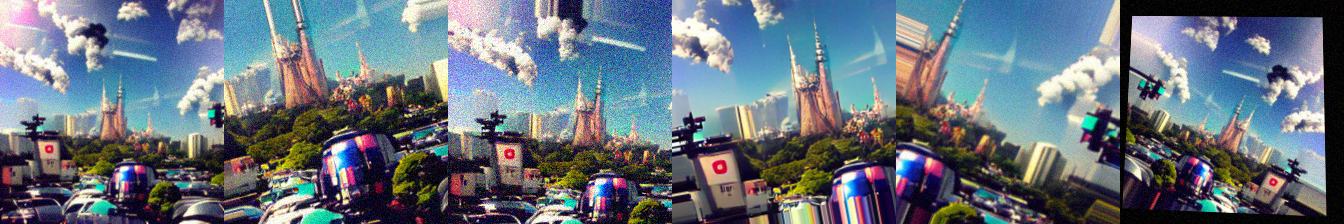}} \\
                \bottomrule
            \end{tabular}
        \end{adjustbox}
        \caption{Prompts closer to each prompt, in baseline (Mean)}
    \end{subfigure}\hfill
    \removespaceovercaption
    \caption{The cutouts (patches) from generated images in Figure~\ref{fig:two-prompts-generated-images}, for both our proposed method (OT) and the baseline. We show in (a) and (b) the sampled patches.
    Then in (c) and (d) we group these patches by the closer (measure by $D$) prompt they are to. Due to space constraints, we only show the number of each group and six patches that are mostly closet.}
    \label{fig:two-prompts-cutouts}
    \removespacebottom
\end{figure*}

\begin{figure*}[hbt!]
    \centering
    \newcommand\subwidth{0.40\textwidth}
    \hspace*{\fill}
    \begin{subfigure}[t]{\subwidth}
        \includegraphics[width=\textwidth]{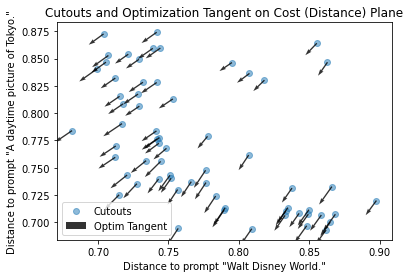}
        \caption{Our method (OT)}
    \end{subfigure}
    \hspace*{\fill}
    \begin{subfigure}[t]{\subwidth}
        \includegraphics[width=\textwidth]{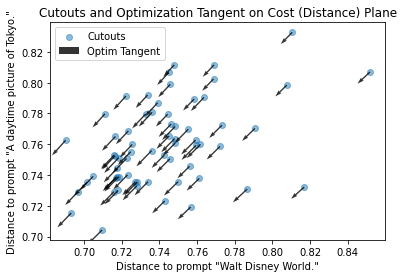}
        \caption{Baseline (Mean)}
    \end{subfigure}
    \hspace*{\fill}
    \removespaceovercaption
    \caption{Tangent after pushforward of the gradients on each patch (cutout) in the embedding space to the cost plane.
    Each blue dot is a patch (cutout),  and intuitively, its coordinate shows the distance to one of two prompts, while its arrow shows the force of gradient that pushes it towards the prompts.
    On the left side, in our method the force of gradient pushes patches to prompts with different ``mix ratio'', promoting the intrinsic diversity in the generated image from which patches are sampled.
    On the right side, in the baseline all patches are pushed for the same mix of prompts, thus leading to less diversity.
    Formally, the exact form and motivation for the tangent could be found mathematically in Equation~\ref{eq:w_i} and its discussions.}
    \label{fig:two-prompts-pushforward}
    \removespacebottom
\end{figure*}

\section{Methodology}
Our motivation comes from the concern arising from using an averaged loss $\mathrm{Mean}_D$.
By focusing on means, all sampled patches are encouraged to move uniformly to the mean of all prompts.
This undermines the very motivation of introducing multiple prompts, which is to allow artists to obtain spatial diversity in the generated images, with various areas reflecting the diversity prompts.
Furthermore, taking the mean in the embedding space introduces gradients in unwanted directions.
Since the locations in the embedding space are associated with semantics, doing so may introduce uncontrollable, redundant semantics.
To make things worse, the mean arithmetic effectively assumes an Euclidean space,
which is inconsistent to the CLIP model that is trained with cosine distance in the embedding space.

To address these issues, it is possible to devise an arithmetic in non-Euclidean Space. 
However, finding a proper choice that works well with the rest of pipeline is not trivial and warrants a separate study.
Instead we propose to eliminate the undesired simplifications brought by mean arithmetics, to replace 
$\mathrm{Mean}_{D}$ in Equation~\ref{eq:f_mean} with an optimal transport loss,
\begin{equation} \label{eq:our_f_mean}
F =  \mathrm{L}^{\epsilon} (a, b, [D(u_i, v_j)]_{i,j})
\end{equation}
where $a_i = 1/n$ and $b_j = 1/m$,
and the cost matrix $\mathbf{C}$ is populated with pairwise distance $D$ evaluations.
Now, the gradient $\nabla_z F$ reads
\begin{equation} \label{eq:our_nabla_f_mean}
\nabla_z F = \left(
 \sum_{1\leq i \leq n}
 \frac{\partial \mathrm{L}^{\epsilon}}{\partial u_i}
 \frac{\partial u_i}{\partial x_i}
 \frac{\partial x_i}{\partial x}
 \right)
 \frac{\partial x}{\partial z}.
\end{equation}
Comparing with Equation~\ref{eq:nabla_f_mean},
the only different term is $\frac{\partial \mathrm{L}^{\epsilon}}{\partial u_i}$ which is also defined as in Equation~\ref{eq:partial_lc}.
Along with other terms (see Equation~\ref{eq:partial_terms}), all terms are defined, and thus we know that $\nabla_z F$ is also well-defined and can be used in the iteratively updating of $z$:
\[
z \gets z - \gamma \nabla_{z} F
\]
 
In doing so, the above mentioned issues are solved for the following reasons:
\begin{itemize}
 \item
   \emph{OT Treats different patches differently}.
   As OT matches patches and text prompts, it naturally introduces a distinct treatment of patches according to their distances to text prompts.
   As the patches are randomly sampled, it encourages the intrinsic diversity \emph{inside} a single generated image.
 \item
   \emph{OT does not involve arithmetic in the latent space}.
   OT relies on distances, but does not use averages in embedding spaces. Therefore it does not produce synthetic prompts in embeddings space that may not correspond to semantics.
   Furthermore, OT is agnostic to how distances are defined: any distance, other than cosine distance or geodesic distance, could be used to populate matrix $\mathbf{C}$.
\end{itemize}
\section{Experiments}
 
In this section, we highlight a few  possibilities brought forward by using our methodology when handling multiple text prompts.
Due to the creative nature of text-to-image synthesis, there is no standard measuring stick, such as classification accuracy, to provide a simple comparison between methods.
Nevertheless, we consider a few tasks that can help us gain insight into the novelty, the properties and the behavior of our method. We consider:
 
\begin{itemize}
     \item \emph{Generated Image}. Naturally the foremost task is to show the generated image $x$ with multiple text prompts $t_1,\cdots,t_m$.
     In this task, we focus on whether the generated image represents the text prompts in a way that is distinctive and subjectively recognized by human viewers.
     \item \emph{Patches (Cutouts) from Generated Images}.
     Our method improves the diversity of patches through increasing the correlation between the distribution of randomly sampled patches and multiple text prompts, as we identify as a source of issues from existing practices.
     In this task, we show the patches and organize them by text prompt.
     Formally, we show the $n$ patches $x_1, \cdots, x_n$ sampled from $x$, and group $x_i$ by $j^{*} = \argmin_{j} D(u_i, v_j)$, the closest text prompt in the embedding space.
     \item \emph{Tangent of Patches (Cutouts) on Cost Plane}.
     We identify the issue materialize in the way gradient information is pass from $F$ back to patches, which is ${\partial \mathrm{Mean}_{D}} / {\partial u_i} $ part in Equation~\ref{eq:nabla_f_mean},
     and propose to use
     $\mathrm{L}_{\mathrm{C}}^{\epsilon}$
     such that the
     ${\partial \mathrm{L}_{\mathrm{C}}^{\epsilon}}/{\partial u_i}$
     part in Equation~\ref{eq:our_nabla_f_mean}, is better.

     To quantitatively qualify such property, a few extra deliberations are needed.
     Concretely, we first define
     \begin{equation*}
           \phi (u_i) : \mathbb{R}^d \to \mathbb{R}^m \defeq [D(u_i, v_1), \cdots, D(u_i, v_m)],
     \end{equation*}
     which is by definition a differentiable mapping from the aforementioned embedding space $\mathbb{R}^d$ to $\mathbb{R}^m $, a $m$-d space of distances to prompts where the $j$-th element is the distance to prompt $j$.
     As ${\partial \mathrm{L}_{\mathrm{C}}^{\epsilon}}/{\partial u_i} \in T_{u_i}$ (the tangent space of $\mathbb{R}^d$ at $u_i$), the pushforward by $\phi$ at $u_i$ is defined as
     $d \phi : T_{u_i} \mathbb{R}^d \to T_{\phi(u_i)} \mathbb{R}^m$ such that when applied to the gradient,
     \begin{equation}\label{eq:w_i}
           w_i = d \phi ({\partial \mathrm{L}_{\mathrm{C}}^{\epsilon}}/{\partial u_i})
     \end{equation}
     is in the tangent space of $\mathbb{R}^m$.
     Intuitively, $w_i$ is a $m$-dimensional vector whose $j$-th element denotes the component of gradient that moves the $i$-th patch towards the $j$-th text prompt.
\end{itemize}
 
\subsection{Comparing our Method with the Baseline for Two Prompts Setting}
 
In this experiment, we focus on a scenario with $M = 2$ prompts, ``\emph{Walt Disney World.}'' and ``\emph{daytime picture of Tokyo.}''
We compare two models, our proposed approach with Optimal Transport (Equation~\ref{eq:our_f_mean}) and the baseline using Mean (Equation~\ref{eq:f_mean}),
with the purpose of investigating the behavior of these methods and the difference made by our approach.
We keep all other configurations the same.
Namely, we use a pre-trained VQGAN~\cite{esser2021taming} on Imagenet dataset, $N=64$ randomly sampled patch, and $1000$ iterations of updating $z$.
We organize the conducted tasks as explained before.
 
 \begin{itemize}
    
    \setlength\itemsep{1em} 
  
     \item \emph{Generated Image} and \emph{Patches (Cutouts) from it}.
     In Figure~\ref{fig:two-prompts-generated-images} we show the generated image from both methods.
     Also in Figure~\ref{fig:two-prompts-cutouts} we show the patches (cutouts) sampled from the generated images at the end of all iterations.
 
     We observe that OT helps generate images where patches (cutouts) are more balanced (\textbf{36/28} vs \textbf{49/15}).
     Furthermore, OT's results are more diverse for two prompts.
     For OT, patches close to ``Walt Disney World.'' are more like close-ups and patches close to
     ``A daytime picture of Tokyo.'' are mostly zoomed-out.
     As patches are randomly done, it reflects the intrinsic property of generated images.
 
     \item \emph{Tangent of Patches (Cutouts) on Cost Plane}.
     We push-forward gradients on the patches' embedding space to this cost plane, as explained in Equation~\ref{eq:w_i}, and show the results in Figure~\ref{fig:two-prompts-pushforward}.
     We observe that our method using OT clearly shows that the positions in the cost plane reveal negative correlation, which means that different parts of the generated images are successfully encouraged to provide contribution to the similarities to different promoters.
     This is cross-verified by the ``fan out'' of tangents pushed forward to the cost plane, which shows the divergent gradients providing patch-specific directions in updating.
     In contrast, baseline methods are simply learning to be the mean of two prompts' embeddings, as the tangents show the uniformed gradient direction which does not distinguish between different prompts.
\end{itemize}
 
\begin{figure*}[hbt!]
    \centering
    
    \newcommand\subwidth{0.23\textwidth}
    
    \hspace*{\fill}
    \begin{subfigure}[t]{\subwidth}
        \includegraphics[width=\textwidth]{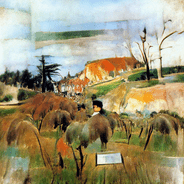}
        \caption{P0: Impressionism / Edgar Degas/ Landscape at Valery-sur-Somme}
    \end{subfigure}
    \hspace*{\fill}
    \begin{subfigure}[t]{\subwidth}
        \includegraphics[width=\textwidth]{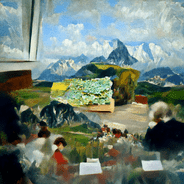}
        \caption{P1: Impressionism Laszlo Mednyanszky/ Landscape in the Alps (View from the Rax)}
    \end{subfigure}
    \hspace*{\fill}
    \begin{subfigure}[t]{\subwidth}
        \includegraphics[width=\textwidth]{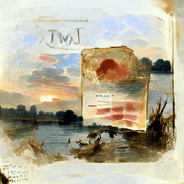}
        \caption{P2: Romanticism / J.M.W. Turner/ The Lake, Petworth, Sunset; Sample Study}
    \end{subfigure}
    \hspace*{\fill}
    
    \hspace*{\fill}
    \begin{subfigure}[t]{\subwidth}
        \includegraphics[width=\textwidth]{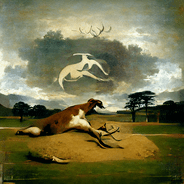}
        \caption{P3: Romanticism / George Stubbs/ Hound Coursing a Stag}
    \end{subfigure}
    \hspace*{\fill}
    \begin{subfigure}[t]{\subwidth}
        \includegraphics[width=\textwidth]{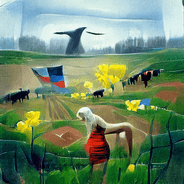}
        \caption{P4: Realism / Alexey Venetsianov/ In the Fields. Spring}
    \end{subfigure}
    \hspace*{\fill}
    \begin{subfigure}[t]{\subwidth}
        \includegraphics[width=\textwidth]{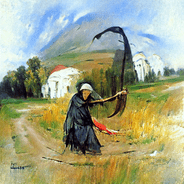}
        \caption{P5: Realism / Alexey Venetsianov/ A Peasant Woman with Scythe and Rake}
    \end{subfigure}
    \hspace*{\fill}
    
    \vspace{0.10cm}
    \noindent\makebox[\linewidth]{\rule{\textwidth}{0.5pt}}
    \vspace{0.01cm}
    
    \hspace*{\fill}
    \begin{subfigure}[t]{\subwidth}
        \includegraphics[width=\textwidth]{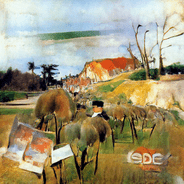}
        \caption{P0}
    \end{subfigure}
    \hspace*{\fill}
    \begin{subfigure}[t]{\subwidth}
        \includegraphics[width=\textwidth]{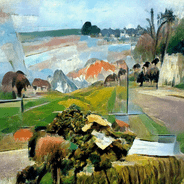}
        \caption{P0 + P1}
    \end{subfigure}
    \hspace*{\fill}
    \begin{subfigure}[t]{\subwidth}
        \includegraphics[width=\textwidth]{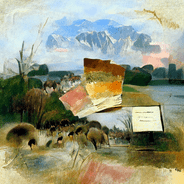}
        \caption{P0 + P1 + P2}
    \end{subfigure}
    \hspace*{\fill}
    
    \hspace*{\fill}
    \begin{subfigure}[t]{\subwidth}
        \includegraphics[width=\textwidth]{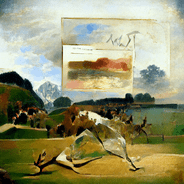}
        \caption{P0 + P1 + P2 + P3}
    \end{subfigure}
    \hspace*{\fill}
    \begin{subfigure}[t]{\subwidth}
        \includegraphics[width=\textwidth]{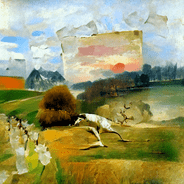}
        \caption{P0 + P1 + P2 + P3 + P4}
    \end{subfigure}
    \hspace*{\fill}
    \begin{subfigure}[t]{\subwidth}
        \includegraphics[width=\textwidth]{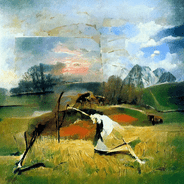}
        \caption{P0 + P1 + P2 + P3 + P4 + P5}
    \end{subfigure}
    \hspace*{\fill}
    
    \caption{The generated images from multiple (6) prompts, labeled P0 to P5. 
    (a) - (f): The first group of 6 images are generated using each one prompt respectively, as a controlling group.
    (g) - (i): The second group of 6 images are the generated images with multiple (1 to 6) prompts respectively from our proposed method,
    each one of which using a combination of multiple problems specified in the caption.}
    \label{fig:ot-multiple-prompts}
    
    \end{figure*}

\begin{figure*}[hbt!]
    \centering
    
    \newcommand\subwidth{0.26\textwidth}
    
    \hspace*{\fill}
    \begin{subfigure}[t]{0.25\textwidth}
        \raisebox{0.0cm}{ 
            \includegraphics[trim=60px 0 106px 20px, clip, width=\textwidth]{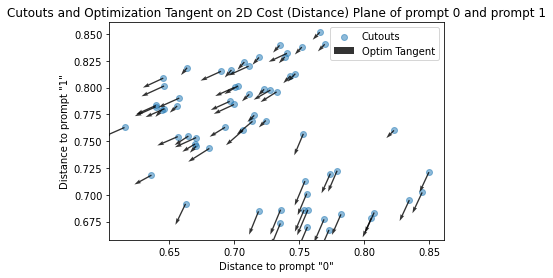}
        }
        \caption{P0 and P1}
    \end{subfigure}
    \hspace*{\fill}
    \begin{subfigure}[t]{0.25\textwidth}
        \raisebox{0.0cm}{
            \includegraphics[trim=60px 0 106px 20px, clip, width=\textwidth]{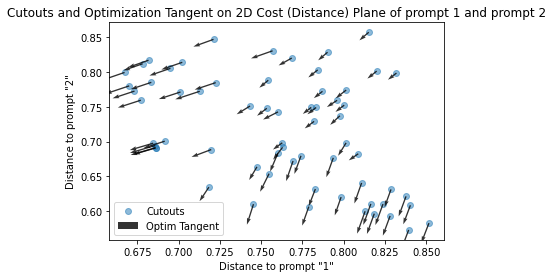}
        }
        \caption{P1 and P2}
    \end{subfigure}
    \hspace*{\fill}
    \begin{subfigure}[t]{0.20\textwidth}
        \includegraphics[width=\textwidth]{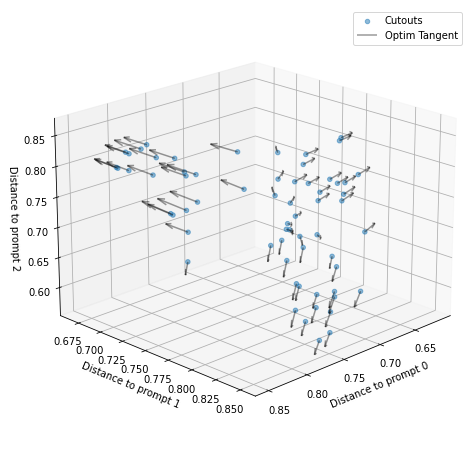}
        \caption{P0, P1 and P2}
    \end{subfigure}
    \hspace*{\fill}
    
    \vspace{0.25cm}
    \noindent\makebox[\linewidth]{\rule{\textwidth}{0.5pt}}
    \vspace{0.00cm}
    
    \hspace*{\fill}
    \renewcommand\subwidth{0.24\textwidth}
    \begin{subfigure}[t]{\subwidth}
        \raisebox{0cm}{
            \includegraphics[trim=60px 0 106px 20px, clip, width=\textwidth]{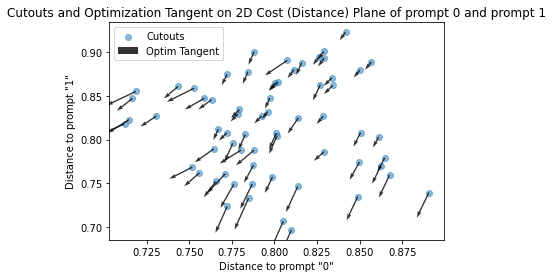}
        }
        \caption{P0 and P1}
    \end{subfigure}
    \hspace*{\fill}
    \begin{subfigure}[t]{\subwidth}
        \raisebox{0cm}{
            \includegraphics[trim=60px 0 106px 20px, clip, width=\textwidth]{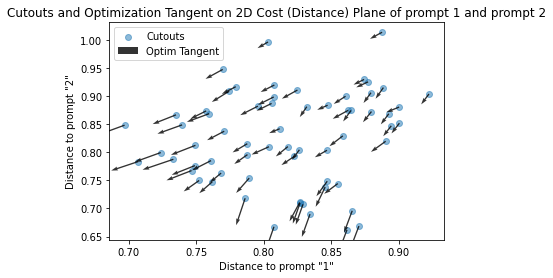}
        }
        \caption{P1 and P2}
    \end{subfigure}
    \hspace*{\fill}
    \begin{subfigure}[t]{\subwidth}
        \raisebox{0cm}{
            \includegraphics[trim=60px 0 106px 20px, clip, width=\textwidth]{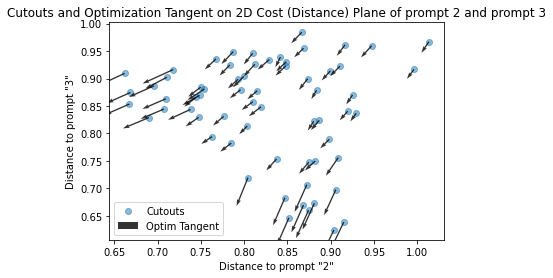}
        }
        \caption{P2 and P3}
    \end{subfigure}
    \hspace*{\fill}
    \begin{subfigure}[t]{\subwidth}
        \raisebox{0cm}{
            \includegraphics[trim=60px 0 106px 20px, clip, width=\textwidth]{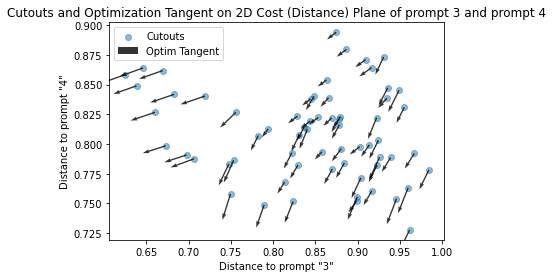}
        }
        \caption{P3 and P4}
    \end{subfigure}
    \hspace*{\fill}
    
    \renewcommand\subwidth{0.19\textwidth}
    \hspace*{\fill}
    \begin{subfigure}[t]{\subwidth}
        \includegraphics[width=\textwidth]{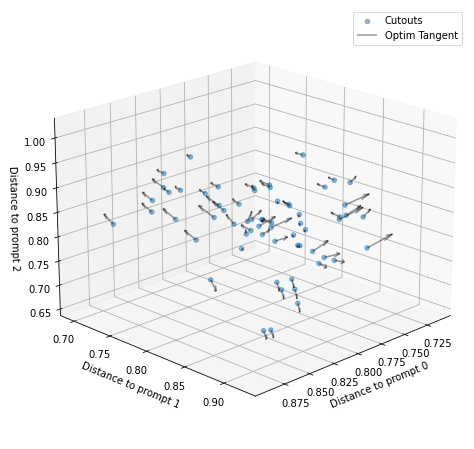}
        \caption{P0, P1 and P2}
    \end{subfigure}
    \hspace*{\fill}
    \begin{subfigure}[t]{\subwidth}
        \includegraphics[width=\textwidth]{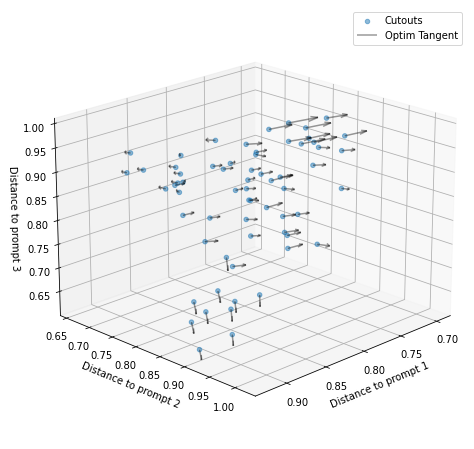}
        \caption{P1, P2 and P3}
    \end{subfigure}
    \hspace*{\fill}
    \begin{subfigure}[t]{\subwidth}
        \includegraphics[width=\textwidth]{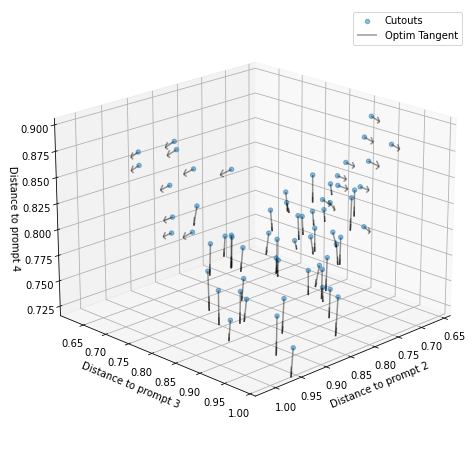}
        \caption{P2, P3 and P4}
    \end{subfigure}
    \hspace*{\fill}

    \removespaceovercaption
    \caption{Tangent, representing the gradients on patches (cutouts) after they are pushed forward to Cost Plan. 
    The first group is for the generation with 3 prompts and the second group is for the generation with 6 prompts, showing in 2D and 3D slices.
    }
    \label{fig:ot-multiple-prompts-part2}
    \removespacebottom
\end{figure*}

\subsection{Our Method's Behavior with Multiple Prompts}
 
Having comparing our OT-based method with the baseline on the two prompts setting, we shift our focus to the scenario where our method is applied to multiple prompts.
As this is we designed our method to expose fine differentiation among prompts, it becomes interesting to investigate such behavior when the number of prompts increases.
In doing so, we consider totally $M=6$ prompts, numbered from P0 to P5:
\begin{itemize}[leftmargin=4.0mm,label=$\ast$]
     \item P0: Impressionism / Edgar Degas/ Landscape at Valery-sur-Somme
     \item P1: Impressionism Laszlo Mednyanszky/ Landscape in the Alps (View from the Rax)
     \item P2: Romanticism / J.M.W. Turner/ The Lake, Petworth, Sunset; Sample Study
     \item P3: Romanticism / George Stubbs/ Hound Coursing a Stag
     \item P4: Realism / Alexey Venetsianov/ In the Fields. Spring
     \item P5: Realism / Alexey Venetsianov/ A Peasant Woman with Scythe and Rake
\end{itemize}
and as the prompts suggest, we use a pre-trained VQGAN on WikiArt dataset consisting mostly of paintings.
The purpose is to both show that our method could be applied to generative models trained from different genre data, and also that the painting allows easier qualitative comparison of both objects and artistic styles.
As the same setting mentioned above, $N=64$ randomly sampled patch, and $1000$ iterations of updating are used. We conduct tasks as explained before.

\begin{itemize}
     \item \emph{Generated Image}.
     In Figure~\ref{fig:ot-multiple-prompts}, we show in the first group the generated images corresponding to these prompts individually,  and in the second group the generated images by combining prompts using our proposed method.
     We observe that our method is capable of composing the instructions from several prompts, in terms of styles and objects, into the same canvas.
 
     \item \emph{Tangent of Patches (Cutouts) on Cost Plane}.
     In Figure~\ref{fig:ot-multiple-prompts-part2}, we show that the good behavior on tangent remains even for multiple prompts.
     This means that our method is capable of guiding generating images that are diverse in its contents w.r.t. multiple prompts.
\end{itemize}

\section{Conclusion and Future Work}
 
In this paper we discuss the problem in dealing with multiple text prompts in the setting of text-driven image generation for computational creativity setting. 
We then propose to address the issue using OT (Optimal Transport) between sampled patches in the generated image and multiple text prompts, and show its theoretical motivation and quantitative and qualitative empirical results highlighting the advantage brought by our proposed method.
 
One of the advantages in our method is that it is in theory orthogonal to other parts in the whole text driven image generation pipeline, as we show primarily that it works for VQGAN trained on several datasets.
We envision that future work would investigate leveraging our proposed method to other drastically different forms of generative method, such as diffusion models.
Another possible future direction may principally study the combination of optimal transport and adaptive sampling where in our proposed work only random sampling is used for simplicity.


\bibliographystyle{iccc}
\bibliography{reference}

\begin{thebibliography}{}

\bibitem[\protect\citeauthoryear{Brown \bgroup et al.\egroup
  }{2020}]{brown2020language}
Brown, T.~B.; Mann, B.; Ryder, N.; Subbiah, M.; Kaplan, J.; Dhariwal, P.;
  Neelakantan, A.; Shyam, P.; Sastry, G.; Askell, A.; et~al.
\newblock 2020.
\newblock Language models are few-shot learners.
\newblock {\em arXiv preprint arXiv:2005.14165}.

\bibitem[\protect\citeauthoryear{Burton-King}{2021}]{samburtonking2021introduction}
Burton-King, S.
\newblock 2021.
\newblock Introduction to vqgan+clip.
\newblock https://bit.ly/3rcedh4.

\bibitem[\protect\citeauthoryear{Crowson}{2021}]{crowson2021clip}
Crowson, K.
\newblock 2021.
\newblock Clip guided diffusion.

\bibitem[\protect\citeauthoryear{Cuturi \bgroup et al.\egroup
  }{2022}]{cuturi2021ott}
Cuturi, M.; Meng-Papaxanthos, L.; Tian, Y.; Bunne, C.; Davis, G.; and Teboul,
  O.
\newblock 2022.
\newblock Optimal transport tools (ott): A jax toolbox for all things
  wasserstein.
\newblock {\em arXiv preprint arXiv:2201.12324}.

\bibitem[\protect\citeauthoryear{Cuturi}{2013}]{cuturi2013sinkhorn}
Cuturi, M.
\newblock 2013.
\newblock Sinkhorn distances: Lightspeed computation of optimal transport.
\newblock {\em Advances in neural information processing systems}
  26:2292--2300.

\bibitem[\protect\citeauthoryear{Danskin}{1966}]{danskin1966theory}
Danskin, J.~M.
\newblock 1966.
\newblock The theory of max-min, with applications.
\newblock {\em SIAM Journal on Applied Mathematics} 14(4):641--664.

\bibitem[\protect\citeauthoryear{Dhariwal and
  Nichol}{2021}]{dhariwal2021diffusion}
Dhariwal, P., and Nichol, A.
\newblock 2021.
\newblock Diffusion models beat gans on image synthesis.
\newblock {\em arXiv preprint arXiv:2105.05233}.

\bibitem[\protect\citeauthoryear{Esser, Rombach, and
  Ommer}{2021}]{esser2021taming}
Esser, P.; Rombach, R.; and Ommer, B.
\newblock 2021.
\newblock Taming transformers for high-resolution image synthesis.
\newblock In {\em Proceedings of the IEEE/CVF Conference on Computer Vision and
  Pattern Recognition},  12873--12883.

\bibitem[\protect\citeauthoryear{Ho, Jain, and Abbeel}{2020}]{ho2020denoising}
Ho, J.; Jain, A.; and Abbeel, P.
\newblock 2020.
\newblock Denoising diffusion probabilistic models.
\newblock {\em arXiv preprint arXiv:2006.11239}.

\bibitem[\protect\citeauthoryear{Jin \bgroup et al.\egroup
  }{2017}]{jin2017towards}
Jin, Y.; Zhang, J.; Li, M.; Tian, Y.; Zhu, H.; and Fang, Z.
\newblock 2017.
\newblock Towards the automatic anime characters creation with generative
  adversarial networks.
\newblock {\em arXiv preprint arXiv:1708.05509}.

\bibitem[\protect\citeauthoryear{Kantorovich}{1942}]{kantorovich1942transfer}
Kantorovich, L.
\newblock 1942.
\newblock On the transfer of masses (in russian).
\newblock In {\em Doklady Akademii Nauk},  227--229.

\bibitem[\protect\citeauthoryear{Karras \bgroup et al.\egroup
  }{2020}]{karras2020analyzing}
Karras, T.; Laine, S.; Aittala, M.; Hellsten, J.; Lehtinen, J.; and Aila, T.
\newblock 2020.
\newblock Analyzing and improving the image quality of stylegan.
\newblock In {\em Proceedings of the IEEE/CVF Conference on Computer Vision and
  Pattern Recognition},  8110--8119.

\bibitem[\protect\citeauthoryear{Karras \bgroup et al.\egroup
  }{2021}]{karras2021alias}
Karras, T.; Aittala, M.; Laine, S.; H{\"a}rk{\"o}nen, E.; Hellsten, J.;
  Lehtinen, J.; and Aila, T.
\newblock 2021.
\newblock Alias-free generative adversarial networks.
\newblock {\em Advances in Neural Information Processing Systems} 34.

\bibitem[\protect\citeauthoryear{Karras, Laine, and
  Aila}{2019}]{karras2019style}
Karras, T.; Laine, S.; and Aila, T.
\newblock 2019.
\newblock A style-based generator architecture for generative adversarial
  networks.
\newblock In {\em Proceedings of the IEEE/CVF Conference on Computer Vision and
  Pattern Recognition},  4401--4410.

\bibitem[\protect\citeauthoryear{Kim and Ye}{2021}]{kim2021diffusionclip}
Kim, G., and Ye, J.~C.
\newblock 2021.
\newblock Diffusionclip: Text-guided image manipulation using diffusion models.
\newblock {\em arXiv preprint arXiv:2110.02711}.

\bibitem[\protect\citeauthoryear{Krantz and Parks}{2002}]{krantz2002implicit}
Krantz, S.~G., and Parks, H.~R.
\newblock 2002.
\newblock {\em The implicit function theorem: history, theory, and
  applications}.
\newblock Springer Science \& Business Media.

\bibitem[\protect\citeauthoryear{Li \bgroup et al.\egroup
  }{2017}]{li2017learning}
Li, A.; Jabri, A.; Joulin, A.; and van~der Maaten, L.
\newblock 2017.
\newblock Learning visual n-grams from web data.
\newblock In {\em Proceedings of the IEEE International Conference on Computer
  Vision},  4183--4192.

\bibitem[\protect\citeauthoryear{Li, Jin, and Zhu}{2021}]{li2021surrogate}
Li, M.; Jin, Y.; and Zhu, H.
\newblock 2021.
\newblock Surrogate gradient field for latent space manipulation.
\newblock In {\em Proceedings of the IEEE/CVF Conference on Computer Vision and
  Pattern Recognition},  6529--6538.

\bibitem[\protect\citeauthoryear{Murdock}{}]{murdock}
Murdock, R.
\newblock @advadnoun.
\newblock https://twitter.com/advadnoun.

\bibitem[\protect\citeauthoryear{Murdock}{2021a}]{wang2021bigsleep}
Murdock, R.
\newblock 2021a.
\newblock Big sleep: A simple command line tool for text to image generation,
  using openai's clip and a biggan.

\bibitem[\protect\citeauthoryear{Murdock}{2021b}]{wang2021deepdaze}
Murdock, R.
\newblock 2021b.
\newblock Deep daze: A simple command line tool for text to image generation
  using openai's clip and siren (implicit neural representation network).

\bibitem[\protect\citeauthoryear{Nichol and
  Dhariwal}{2021}]{nichol2021improved}
Nichol, A., and Dhariwal, P.
\newblock 2021.
\newblock Improved denoising diffusion probabilistic models.
\newblock {\em arXiv preprint arXiv:2102.09672}.

\bibitem[\protect\citeauthoryear{Nichol \bgroup et al.\egroup
  }{2021}]{nichol2021glide}
Nichol, A.; Dhariwal, P.; Ramesh, A.; Shyam, P.; Mishkin, P.; McGrew, B.;
  Sutskever, I.; and Chen, M.
\newblock 2021.
\newblock Glide: Towards photorealistic image generation and editing with
  text-guided diffusion models.
\newblock {\em arXiv preprint arXiv:2112.10741}.

\bibitem[\protect\citeauthoryear{Patashnik \bgroup et al.\egroup
  }{2021}]{patashnik2021styleclip}
Patashnik, O.; Wu, Z.; Shechtman, E.; Cohen-Or, D.; and Lischinski, D.
\newblock 2021.
\newblock Styleclip: Text-driven manipulation of stylegan imagery.
\newblock In {\em Proceedings of the IEEE/CVF International Conference on
  Computer Vision},  2085--2094.

\bibitem[\protect\citeauthoryear{Peyr{\'e} and
  Cuturi}{2019}]{peyre2019computational}
Peyr{\'e}, G., and Cuturi, M.
\newblock 2019.
\newblock Computational optimal transport: With applications to data science.
\newblock {\em Foundations and Trends{\textregistered} in Machine Learning}
  11(5-6):355--607.

\bibitem[\protect\citeauthoryear{Radford \bgroup et al.\egroup
  }{2021}]{radford2021learning}
Radford, A.; Kim, J.~W.; Hallacy, C.; Ramesh, A.; Goh, G.; Agarwal, S.; Sastry,
  G.; Askell, A.; Mishkin, P.; Clark, J.; et~al.
\newblock 2021.
\newblock Learning transferable visual models from natural language
  supervision.
\newblock {\em arXiv preprint arXiv:2103.00020}.

\bibitem[\protect\citeauthoryear{Raffel \bgroup et al.\egroup
  }{2019}]{raffel2019exploring}
Raffel, C.; Shazeer, N.; Roberts, A.; Lee, K.; Narang, S.; Matena, M.; Zhou,
  Y.; Li, W.; and Liu, P.~J.
\newblock 2019.
\newblock Exploring the limits of transfer learning with a unified text-to-text
  transformer.
\newblock {\em arXiv preprint arXiv:1910.10683}.

\bibitem[\protect\citeauthoryear{S{\'e}journ{\'e} \bgroup et al.\egroup
  }{2019}]{sejourne2019sinkhorn}
S{\'e}journ{\'e}, T.; Feydy, J.; Vialard, F.-X.; Trouv{\'e}, A.; and Peyr{\'e},
  G.
\newblock 2019.
\newblock Sinkhorn divergences for unbalanced optimal transport.
\newblock {\em arXiv preprint arXiv:1910.12958}.

\bibitem[\protect\citeauthoryear{Shorten and
  Khoshgoftaar}{2019}]{shorten2019survey}
Shorten, C., and Khoshgoftaar, T.~M.
\newblock 2019.
\newblock A survey on image data augmentation for deep learning.
\newblock {\em Journal of Big Data} 6(1):1--48.

\bibitem[\protect\citeauthoryear{Snell}{2021}]{snell2021}
Snell, C.
\newblock 2021.
\newblock Alien dreams: An emerging art scene.
\newblock https://ml.berkeley.edu/blog/posts/clip-art/.

\bibitem[\protect\citeauthoryear{Song, Meng, and
  Ermon}{2020}]{song2020denoising}
Song, J.; Meng, C.; and Ermon, S.
\newblock 2020.
\newblock Denoising diffusion implicit models.
\newblock {\em arXiv preprint arXiv:2010.02502}.

\bibitem[\protect\citeauthoryear{Thomee \bgroup et al.\egroup
  }{2016}]{thomee2016yfcc100m}
Thomee, B.; Shamma, D.~A.; Friedland, G.; Elizalde, B.; Ni, K.; Poland, D.;
  Borth, D.; and Li, L.-J.
\newblock 2016.
\newblock Yfcc100m: The new data in multimedia research.
\newblock {\em Communications of the ACM} 59(2):64--73.

\bibitem[\protect\citeauthoryear{van~den Oord, Vinyals, and
  Kavukcuoglu}{2017}]{van2017neural}
van~den Oord, A.; Vinyals, O.; and Kavukcuoglu, K.
\newblock 2017.
\newblock Neural discrete representation learning.
\newblock In {\em Proceedings of the 31st International Conference on Neural
  Information Processing Systems},  6309--6318.

\end{thebibliography}
\end{document}